\pdfoutput=1
\documentclass[11pt]{article}

\usepackage[]{emnlp2021}

\usepackage[T1]{fontenc}
\usepackage{amsfonts}
\usepackage[utf8]{inputenc}
\usepackage{microtype}
\usepackage{amsmath}
\usepackage{array}

\usepackage{times}
\usepackage{latexsym}
\usepackage{multirow}
\usepackage{multicol}
\usepackage{color, colortbl}
\usepackage{longtable}
\usepackage{graphicx}
\usepackage{makecell}
\usepackage{booktabs}
\usepackage{color}
\usepackage{subfigure}

\definecolor{Gray}{gray}{0.9}

\newcommand{\finalcopy}[1]{{#1}}

\newcommand{\tabincell}[2]{\begin{tabular}{@{}#1@{}}#2\end{tabular}} 

\title{Benchmarking Commonsense Knowledge Base Population with an Effective Evaluation Dataset}

\author{Tianqing Fang$^{1}$\thanks{\quad Equal Contribution} , Weiqi Wang$^{1*}$, Sehyun Choi$^1$,\\
{\bf Shibo Hao$^2$,} {\bf Hongming Zhang$^1$,} {\bf Yangqiu Song$^1$,}  {\bf \& Bin He$^3$} \\
$^{1}$Department of Computer Science and Engineering, HKUST \\
$^{2}$School of Electronics Engineering and Computer Science, Peking University \\
$^{3}$Huawei Noah's Ark Lab \\
\texttt{tfangaa@cse.ust.hk}, \texttt{\{wwangbw,schoiaj\}@connect.ust.hk},\\
\texttt{haoshibo@pku.edu.cn}, \texttt{\{hzhangal,yqsong\}@cse.ust.hk},\\ \texttt{hebin.nlp@huawei.com}
\\
}

\begin{document}

\maketitle

\begin{abstract}

Reasoning over commonsense knowledge bases (CSKBs) whose elements are in the form of free-text is an important yet hard task in NLP. While {\it CSKB completion} only fills the missing links within the domain of the CSKB, {\it CSKB population} is alternatively proposed with the goal of reasoning unseen assertions from external resources.
In this task, CSKBs are grounded to a large-scale eventuality (activity, state, and event) graph to discriminate whether novel triples from the eventuality graph are plausible or not.
However, existing evaluations on the population task are either not accurate (automatic evaluation with randomly sampled negative examples) or of small scale (human annotation). 
In this paper, we benchmark the CSKB population task with a new large-scale dataset by first aligning four popular CSKBs, and then presenting a high-quality human-annotated evaluation set to probe neural models' commonsense reasoning ability.
We also propose a novel inductive commonsense reasoning model that reasons over graphs.
Experimental results show that generalizing commonsense reasoning on unseen assertions is inherently a hard task. Models achieving high accuracy during training perform poorly on the evaluation set, with a large gap between human performance. 
\finalcopy{Codes and data are available at \url{https://github.com/HKUST-KnowComp/CSKB-Population}.}
\end{abstract}

\section{Introduction}

\begin{figure}[t]
    \centering
    \includegraphics[width=\linewidth]{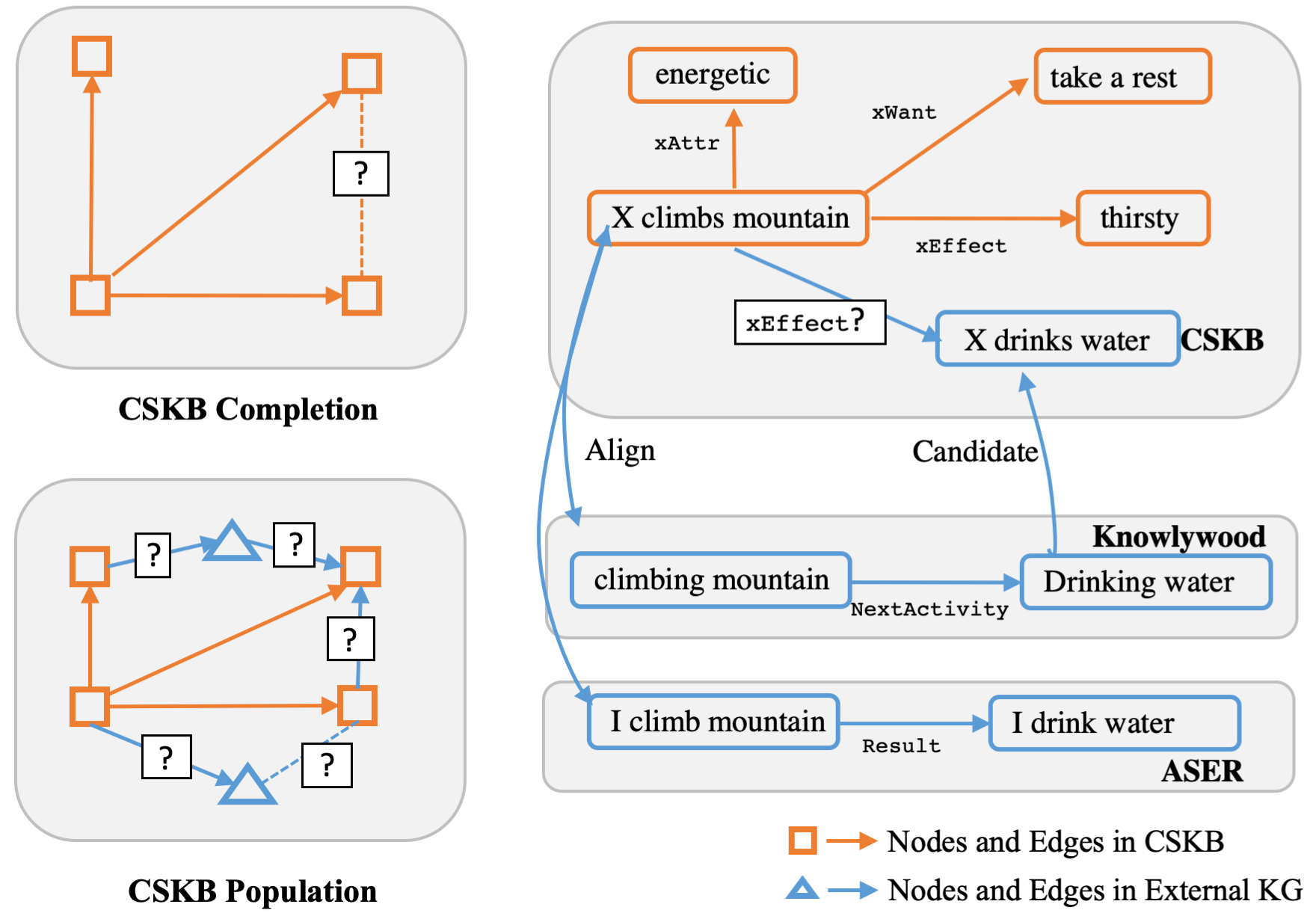}
    \caption{ Comparison between CSKB completion and population. An example of aligning eventuality graph as candidate commonsense knowledge triples is also provided.}
    \label{fig:introduction}
\end{figure}

Commonsense reasoning is one of the core problems in the field of artificial intelligence. Throughout the development in computational commonsense, commonsense knowledge bases (CSKB) \cite{speer2017conceptnet,sap2019atomic} are constructed to enhance models' reasoning ability. 
As human-annotated CSKBs are far from complete due to the scale of crowd-sourcing, 
reasoning tasks such as {\it CSKB completion}~\cite{li2016commonsense, malaviya2020commonsense, moghimifar2021neuralsymbolic} and {\it population}~\cite{fang2021discos} are proposed to enrich the missing facts.
The CSKB completion task is defined based on the setting of predicting missing links within the CSKB.
On the other hand, the population task grounds commonsense knowledge in CSKBs to large-scale automatically extracted candidates, and requires models to determine whether a candidate triple, \textit{(head, relation, tail)},
is plausible or not, based on the information from both the CSKB and the large number of candidates which essentially form a large-scale graph structure. 
An illustration of the difference between completion and population is shown in Figure~\ref{fig:introduction}.

There are two advantages of the population task. First, the population can not only add links but also nodes to an existing CSKB, while completion can only add links. 
The populated CSKB can also help reduce the \textit{selection bias} problem~\cite{heckman1979sample} from which most machine learning models would suffer, and will benefit a lot of downstream applications such as commonsense generation~\cite{bosselut2019comet}.
\finalcopy{Second, commonsense knowledge is usually implicit knowledge that requires multiple-hop reasoning, while current CSKBs are lacking such complex graph structures.
For example, in ATOMIC~\cite{sap2019atomic}, a human-annotated {\it if-then} commonsense knowledge base among daily events and (mental) states, the average hops between matched heads and tails in ASER, an automatically extracted knowledge base among activities, states, and events based on discourse relationships, is 2.4~\cite{zhang2021aser}. 
Evidence in Section~\ref{section:aligned_graph}  (Table~\ref{table:overall_match_coverage}) also shows similar results for other CSKBs.
However, reasoning solely on existing CSKBs can be viewed as a simple triple classification task without considering complex graph structure (as shown in Table~\ref{table:overall_match_coverage}, the graphs in CSKBs are much sparser).
The population task, which provides a richer graph structure, can explicitly leverage the large-scale corpus to perform commonsense reasoning over multiple hops on the graph.
}

However, there are two major limitations
for the evaluation of the CSKB population task.
First, automatic evaluation metrics, which are based on distinguishing ground truth annotations from automatically sampled negative examples (either a random head or a random tail), are not accurate enough. 
Instead of directly treating the random samples as \textit{negative}, solid human annotations are needed to provide hard labels for commonsense triples.
Second, the human evaluation in the original paper of CSKB population~\cite{fang2021discos} cannot be generally used for benchmarking.
They first populate the CSKB and then asked human annotators to annotate a small subset to check whether the populated results are accurate or not.
A better benchmark should be based on random samples from all candidates and the scale should be large enough to cover diverse events and states. 

To effectively and accurately evaluate CSKB population, 
in this paper, we benchmark CSKB population by firstly proposing a comprehensive dataset aligning four popular CSKBs and a large-scale automatically extracted knowledge graph, and then providing a large-scale human-annotated evaluation set.
Four event-centered CSKBs that cover daily events, namly ConceptNet~\cite{speer2017conceptnet} (the event-related relations are selected), ATOMIC~\cite{sap2019atomic}, ATOMIC$_{20}^{20}$~\cite{hwang2020comet}, and GLUCOSE~\cite{mostafazadeh2020glucose}, are used to constitute the commonsense relations. 
We align the CSKBs together into the same format and ground them to a large-scale eventuality (including activity,  state, and event) knowledge graph, ASER~\cite{zhang2020aser,zhang2021aser}. 
Then, instead of annotating every possible node pair in the graph, which takes an infeasible $O(|V|^2)$ amount of annotation,
we sample a large subset of candidate edges grounded in ASER to annotate.
In total, 31.7K high-quality triples are annotated as the development set and test set. 

To evaluate the commonsense reasoning ability of machine learning models based on our benchmark data, we first propose some models that learn to perform CSKB population inductively over the knowledge graph.
Then we conduct extensive evaluations and analysis of the results to demonstrate that CSKB population is a hard task where models perform poorly on our evaluation set far below human performance. 

We summarize the contributions of the paper as follow:
(1) We provide a novel benchmark for CSKB population over new assertions that cover four human-annotated CSKBs, with a large-scale human-annotated evaluation set. (2) We propose a novel inductive commonsense reasoning model that incorporates both semantics and graph structure. (3) We conduct extensive experiments and evaluations on how different models,  commonsense resources for training, and graph structures may influence the commonsense reasoning results. 

\section{Related Works}

\subsection{Commonsense Knowledge Bases}

Since the proposal of Cyc~\cite{lenat1995cyc} and ConceptNet~\cite{liu2004conceptnet, speer2017conceptnet}, a growing number of large-scale human-annotated CSKBs are developed~\cite{sap2019atomic,BiskZLGC20,SakaguchiBBC20,mostafazadeh2020glucose,DBLP:conf/emnlp/ForbesHSSC20socialchem,DBLP:journals/corr/abs-2008-09094,hwang2020comet,ilievski2020cskg}. While ConceptNet mainly depicts the commonsense relations between entities and only small portion of events, recent important CSKBs have been more devoted to event-centric commonsense knowledge. 
For example, ATOMIC~\cite{sap2019atomic} defines 9 social interaction relations and 
\textasciitilde 880K triples are annotated. 
ATOMIC$_{20}^{20}$~\cite{hwang2020comet} further unifies the relations with ConceptNet, together with several new relations, to form a larger CSKB containing 16 event-related relations. 
Another CSKB is GLUCOSE~\cite{mostafazadeh2020glucose}, which extracts sentences from ROC Stories and defines 10 commonsense dimensions to explores the causes and effects given the base event. 
In this paper, we select ConceptNet, ATOMIC, ATOMIC$_{20}^{20}$, and GLUCOSE to align them together because they are all event-centric and relatively 
\finalcopy{more normalized}
compared to other CSKBs like  SocialChemistry101~\cite{DBLP:conf/emnlp/ForbesHSSC20socialchem}. 


\subsection{Knowledge Base Completion and Population}

Knowledge Base (KB) completion is well studied using knowledge base embedding learned from triples~\cite{bordes2013translating, DBLP:journals/corr/YangYHGD14a, DBLP:conf/iclr/SunDNT19} and graph neural networks with a scoring function decoder~\cite{shang2019end}. Pretrained language models are also applied on such completion task~\cite{yao2019kg, DBLP:journals/corr/abs-2004-14781} where information of knowledge triples is translated into the input to \textsc{Bert}~\cite{devlin2019bert} or RoBERTa~\cite{liu2019roberta}. 
Knowledge base population~\cite{ji2011knowledge} typically includes entity linking~\cite{shen2014entity} and slot filling~\cite{surdeanu2014overview} for conventional KBs, and many relation extraction approaches have been proposed~\cite{RothY02,ChanR10,mintz2009distant,RiedelYM10,HoffmannZLZW11,SurdeanuTNM12,LinSLLS16,ZengLLS17}. \textit{Universal schema} and matrix factorization can also be used to learn latent features of databases and perform population~\cite{riedel2013relation,VergaBSRM16,ToutanovaCPPCG15, DBLP:conf/eacl/McCallumNV17}.

Besides completion tasks on conventional entity-centric KBs like Freebase~\cite{bollacker2008freebase}, completion tasks on CSKBs
are also studied on ConceptNet and ATOMIC. Bi-linear models are used to conduct triple classification on ConceptNet~\cite{li2016commonsense, saito2018commonsense}. Besides, knowledge base embedding models plus \textsc{Bert}-based graph densifier~\cite{malaviya2020commonsense, wang2020inductive} are used to perform link prediction.
For CSKB population, \textsc{Bert} plus GraphSAGE~\cite{hamilton2017inductive} is designed to learn a reasoning model on unseen assertions~\cite{fang2021discos}.

Commonsense knowledge generation, such as COMET~\cite{bosselut2019comet} and LAMA~\cite{PetroniRRLBWM19}, is essentially a CSKB population problem. However, it requires the known heads and relations to acquire more tails so it does not fit our evaluation.
Recently, various prompts are proposed to change the predicate lexicalization~\cite{JiangXAN20,ShinRLWS20,abs-2104-05240} but still how to obtain more legitimate heads for probing remains unclear.
Our work can benefit them by obtaining more training examples, mining more commonsense prompts, as well as getting more potential heads for the generation.



\section{Task Definition}

Denote the source CSKB about events as $\mathcal{C}=\{(h, r, t) | h \in \mathcal{H}, r\in \mathcal{R}, t\in \mathcal{T}\}$, where $\mathcal{H}$, $\mathcal{R}$, and $\mathcal{T}$ are the set of the commonsense heads, relations, and tails. Suppose we have another much larger eventuality (including activity,  state, and event) knowledge graph extracted from texts, denoted as $\mathcal{G}=(\mathcal{V}, \mathcal{E})$, where $\mathcal{V}$ is the set of all vertices and $\mathcal{E}$ is the set of edges. $\mathcal{G}^c$ is the graph acquired by aligning $\mathcal{C}$ and $\mathcal{G}$ into the same format. 
The goal of CSKB population is to learn a scoring function given a candidate triple $(h, r, t)$, \finalcopy{where plausible commonsense triples should be scored higher}. The training of CSKB population can inherit the setting of triple classification, where ground truth examples are from the CSKB $\mathcal{C}$ and negative triples are randomly sampled. In the evaluation phase, the model is required to score the triples from $\mathcal{G}$ that are not included in $\mathcal{C}$ and be compared with human-annotated labels.  

\begin{table}[t]
\small
\centering
\begin{tabular}{l|l}
\toprule
Glucose & ATOMIC Relations \\
\midrule
Dim 1, 6 & xEffect, oEffect \\
Dim 2 & xAttr (``feels''), xIntent (otherwise) \\
Dim 3, 4, 8, 9 & Causes \\
Dim 5, 10 & xWant, oWant \\
Dim 7 & xReact, oReact\\
\bottomrule
\end{tabular}
\caption{The conversion from GLUCOSE relations to ATOMIC$_{20}^{20}$ relations, inherited from~\citet{mostafazadeh2020glucose}.}
\label{table:glucose_edge_convertion}
\end{table}

\begin{table*}[t]
\small
\centering
\begin{tabular}{c|cccc|c}
\toprule
~&\tabincell{c}{ATOMIC\\ (No clause)} & \tabincell{c}{ATOMIC$_{20}^{20}$\\ (4 relations)}&\tabincell{c}{ConceptNet \\(Event-centered)}& GLUCOSE & \# Eventuality\\
\midrule
\# Triples & 449,056 & 124,935 & 10,159 & 117,828 & - \\
\midrule
Knowlywood & 2.63\% & 2.87\% & 16.50\% & 2.96\% & 929,546 \\
ASER & 61.95\% & 38.50\% & 44.94\% & 84.57\% & 52,940,258 \\
\bottomrule
\end{tabular}
\caption{Overlaps between eventuality graphs and commonsense knowledge graphs. We report the proportion of $(h, r, t)$ triples where both the head and tail can be found in the eventuality graph. }
\label{table:event_graph_coverage}
\end{table*}

\section{Dataset Preparation}

\subsection{Selection of CSKBs}\label{sec:select_cskb}

As we aim at exploring commonsense relations among general events, 
we summarize several criteria for selecting CSKBs. 
First, the CSKB should be well symbolically structured to be generalizable. While the nodes in CSKB can inevitably be free-text to represent more diverse semantics, we select the
knowledge resources where format normalization is conducted. 
Second, the commonsense relations are encoded as \textit{(head, relation, tail)} triples.
To this end, among all CSKB resources, we choose the event-related relations in ConceptNet, ATOMIC, ATOMIC$_{20}^{20}$, and GLUCOSE as the final commonsense resources. 
For the event-related relations in ConceptNet, the elements are mostly lemmatized \textit{predicate-object} pairs. In 
ATOMIC and ATOMIC$_{20}^{20}$, the subjects of eventualities are normalized to placeholders ``\textit{PersonX}'' and ``\textit{PersonY}''.
The nodes in GLUCOSE are also normalized and syntactically parsed manually, where human-related pronouns are written as ``\textit{SomeoneA}'' or ``\textit{SomeoneB}'', and object-related pronouns are written as ``\textit{SomethingA}''. Other commonsense resources like SocialChemistry101~\cite{DBLP:conf/emnlp/ForbesHSSC20socialchem} are not selected as they include over loosely-structured events.

For ConceptNet, we select the event-related relations \texttt{Causes} and \texttt{HasSubEvent}, and the triples where nodes are noun phrases are filtered out. For ATOMIC, we restrict the events to be those simple and explicit events that do not contain wildcards and clauses. As ATOMIC$_{20}^{20}$ itself includes the triples in ATOMIC and ConceptNet, to distinguish different relations, we refer to ATOMIC$_{20}^{20}$ as the new event-related relations annotated in \citet{hwang2020comet}, which are \texttt{xReason}, \texttt{HinderedBy}, \texttt{isBefore}, and \texttt{isAfter}.
In the rest of the paper, 
ATOMIC($_{20}^{20}$) means the combination of ATOMIC and the new relations in ATOMIC$_{20}^{20}$.

\subsection{Alignment of CSKBs}\label{sec:cskb_preproc}

To effectively align 
the four CSKBs, we propose best-effort rules to align the formats for both nodes and edges. 
First, for the nodes in each CSKB, we normalize the \textit{person-centric} subjects and objects as ``\textit{PersonX}'', ``\textit{PersonY}'', and ``\textit{PersonZ}'', etc, according to the order of their occurrence, and the \textit{object-centric} subjects and objects as ``\textit{SomethingA}'' and ``\textit{SomethingB}''. Second, to reduce the semantic overlaps of different relations, we aggregate all commonsense relations to the relations defined in ATOMIC($_{20}^{20}$), as it is comprehensive enough to cover the relations in other resources like GLUCOSE, with some simple alignment in Table~\ref{table:glucose_edge_convertion}. 

\noindent \textbf{ConceptNet}. We select \texttt{Causes} and \texttt{HasSubEvent} from ConceptNet to constitute the event-related relations. As heads and tails in ConceptNet don't contain subjects, we add a ``\textit{PersonX}'' in front of the original heads and tails to make them complete eventualities. 

\noindent \textbf{ATOMIC($_{20}^{20}$)}. In ATOMIC and ATOMIC$_{20}^{20}$, heads are structured events with ``\textit{PersonX}'' as subjects, while tails are human-written free-text 
where subjects tend to be missing. We add ``\textit{PersonX}'' for the tails without subjects under \textit{agent}-driven relations, the relations that aim to investigate causes or effects on ``\textit{PersonX}'' himself, and add ``\textit{PersonY}'' for the tails missing subjects under \textit{theme}-driven relations, the relations that investigate commonsense causes or effects on other people like ``\textit{PersonY}'' .

\noindent \textbf{GLUCOSE}. For GLUCOSE, we leverage the parsed and structured version in this study. We replace the personal pronouns ``\textit{SomeoneA}'' and ``\textit{SomeoneB}'' with ``\textit{PersonX}'' and ``\textit{PersonY}'' respectively. For other \textit{object-centric} placeholders like ``\textit{Something}'', we keep them unchanged. The relations in GLUCOSE are then converted to ATOMIC relations according to the conversion rule in the original paper~\cite{mostafazadeh2020glucose}. Moreover, \texttt{gWant}, \texttt{gReact}, and \texttt{gEffect} are the new relations for the triples in GLUCOSE where the subjects are \textit{object-centric}. The prefix ``g'' stands for \textit{general}, to be distinguished from ``x'' (for \textit{PersonX}) and ``o'' (for \textit{PersonY}).

\begin{table*}[t]
\small
\centering
\begin{tabular}{lcccc|cc|cc|cc|cc}
\toprule
\multicolumn{1}{c|}{} &   \multicolumn{4}{c|}{ \multirow{2}{3cm}{ASER$_{norm}$ Coverage}}  & \multicolumn{4}{c|}{ Avg. Degree in ASER$_{norm}$} & \multicolumn{4}{c}{ Avg. Degree in $\mathcal{C}$}  \\ \cline{6-13}
\multicolumn{1}{c|}{~} & ~ & ~ & ~ & ~ & \multicolumn{2}{c|}{In-Degree} & \multicolumn{2}{c|}{Out-Degree}  & \multicolumn{2}{c|}{In-Degree} & \multicolumn{2}{c}{Out-Degree}  \\ \cline{2-13}
\multicolumn{1}{c|}{} & head\scriptsize{(\%)} & tail\scriptsize{(\%)} & edge\scriptsize{(\%)} & \multicolumn{1}{c|}{\#hops}    & head & tail & head & tail &head & tail & head & tail \\ 
\midrule
\multicolumn{1}{c|}{ATOMIC} &  79.76 & 77.11 & 59.32 & 2.57 &  90.9 & 61.3 & 91.2 & 61.6 & 4.2 & 3.4  & 34.6 & 1.5 \\
\multicolumn{1}{c|}{ATOMIC$_{{20}}^{20}$}  & 80.39 & 47.33 & 36.73  & 2.65 &  96.9 & 66.9 & 97.3 & 67.3 & 4.3 & 2.9  & 34.6 & 1.5 \\
\multicolumn{1}{c|}{ConceptNet}  & 77.72 & 54.79 & 43.51 &  2.37  &  210.7 & 88.9 & 211.6 & 88.9 & 15.1 & 8.0  & 26.2 & 4.1\\
\multicolumn{1}{c|}{GLUCOSE}   &91.48 & 91.85 & 81.01 & 2.37  &  224.9 & 246.4 & 226.6 & 248.0 & 7.2 & 7.7  & 6.7 & 5.5 \\ \bottomrule
\end{tabular}
\caption{The overall matching statistics for the four CSKBs. The \textit{edge} column indicates the proportion of edges where their heads and tails can be connected by paths in ASER. 
Average (in and out)-degree on ASER$_{norm}$ and $\mathcal{C}$ for nodes from the CSKBs is also presented. The statistics in $\mathcal{C}$ is different from \cite{malaviya2020commonsense} as we check the degree on the aligned CSKB $\mathcal{C}$ instead of each individual CSKB.
}
\label{table:overall_match_coverage}
\end{table*}

\subsection{Selection of the Eventuality KG}

Taking scale and the diversity of relationships in the KG into account, we select two automatically extracted eventuality knowledge graphs as candidates for the population task,  Knowlywood~\cite{tandon2015knowlywood} and ASER~\cite{zhang2020aser}. 
They both have complex graph structures that are suitable for multiple-hop reasoning.
We first check how much
commonsense knowledge is included in those eventuality graphs to see if it's possible to ground a large proportion of commonsense knowledge triples on the graphs. 
Best-effort alignment rules are designed to align the formats of CSKBs and eventuality KGs.
For Knowlywood, as the patterns are mostly simple \textit{verb-object} pairs, we leverage the \textit{v-o} pairs directly and add a subject in front of the pairs.
For ASER, we aggregate the raw personal pronouns like \textit{he} and \textit{she} to normalized ``\textit{PersonX}''.
As ASER adopts more complicated patterns of defining eventualities, a more detailed pre-process of the alignment between ASER and CSKBs will be illustrated in Section~\ref{sec:aser_preproc}. 
We report the proportion of triples in every CSKB whose head and tail can both be matched to the eventuality graph in Table~\ref{table:event_graph_coverage}. ASER covers a significantly larger proportion of head-tail pairs in the four CSKBs than Knowlywood. 
The reason behind is that on the one hand ASER is of much larger scale, and on the other hand ASER contains eventualities with more complicated structures like \textit{s-v-o-p-o} (\textit{s} for \textit{subject}, \textit{v} for \textit{verb}, \textit{o} for \textit{object}, and \textit{p} for \textit{preposition}), compared with the fact that Knowlywood mostly covers \textit{s-v} or \textit{s-v-o} only.
In the end, we select ASER as the eventuality graph for population.



\subsection{Pre-process of the Eventuality Graph} \label{sec:aser_preproc}

\begin{figure}[t]
    \centering
    \includegraphics[width=0.9\linewidth]{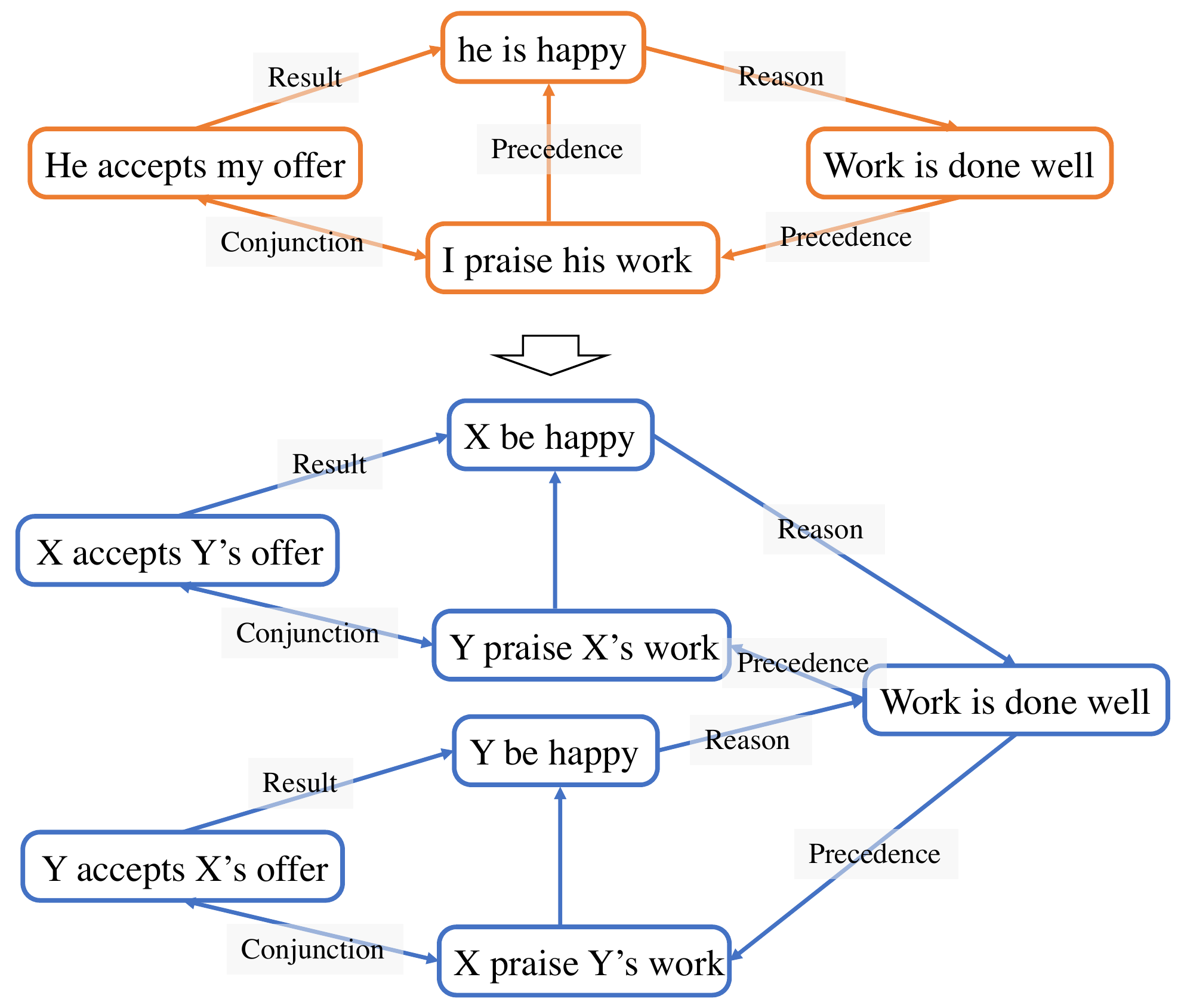}
    \caption{An example of normalizing ASER. The coral nodes and edges are raw data from ASER, and the blue ones are the normalized graph by converting ``he'' and ``she'' to placeholders ``\textit{PersonX}'' and ``\textit{PersonY}''}
    \label{fig:convert_aser}
\end{figure}

We introduce the normalization process of ASER, which converts its knowledge among everyday eventualities into normalized form to be aligned with the CSKBs as discussed in Section~\ref{sec:cskb_preproc}.
Each eventuality in ASER has a subject. 
We consider singular personal pronouns, i.e., ``I'', ``you'', ``he'', ``she'', ``someone'', ``guy'', ``man'', ``woman'', ``somebody'', and replace the concrete personal pronouns in ASER with
normalized formats such as ``\textit{PersonX}'' and ``\textit{PersonY}''. 
Specifically, for an original ASER edge where both the head and tail share the same \textit{person-centric} subject, we replace the subject with ``\textit{PersonX}'' and the subsequent personal pronouns in the two eventualities with ``\textit{PersonY}'' and ``\textit{PersonZ}'' according to the order of the occurrence if exists. 
For the two neighboring eventualities where the subjects are different \textit{person-centric} pronouns, we replace one with ``\textit{PersonX}'' and the other with ``\textit{PersonY}''. 
In addition, to preserve the complex graph structure in ASER, for all the converted edges, we duplicate them by replacing the ``\textit{PersonX}'' in it with ``\textit{PersonY}'', and ``\textit{PersonY}'' with ``\textit{PersonX}'', to preserve the sub-structure in ASER as much as possible. 
An illustration of the converting process is shown in Figure~\ref{fig:convert_aser}. The normalized version of ASER is denoted as ASER$_{norm}$.

%
%

\subsection{The Aligned Graph $\mathcal{G}^c$}\label{section:aligned_graph}

With the pre-process
in Section~\ref{sec:cskb_preproc} and~\ref{sec:aser_preproc}, we can successfully align the CSKBs and ASER together in the same format. To demonstrate ASER's coverage on the knowledge in CSKBs, we present the proportion of heads, tails, and edges that can be found in the ASER$_{norm}$ via exact string match in Table~\ref{table:overall_match_coverage}. 
For edges, we report the proportion of edges where the corresponding heads and tails can be connected by a path in ASER.
We also report the average shortest path length in ASER for those matched edges from the CSKB in the \#hops column, showing that ASER can entail such commonsense knowledge within several hops of path reasoning, which builds the foundation of commonsense reasoning on ASER. 
In addition, the average degree in $\mathcal{G}^c$ and $\mathcal{C}$ for heads and tails from each CSKB is also presented in the table.
The total number of triples for each relation in the CSKBs is presented in Table~\ref{table:cskb_rel_stat}. 
There are 18 commonsense relations in total for CSKBs and 15 relations in ASER.
More detailed descriptions and examples of the unification are presented in the Appendix (Table~\ref{table:cskb_rel_def}, \ref{table:aser_rel_def}, and~\ref{table:alignment_example}). 

\subsection{Evaluation Set Preparation}\label{section:eval_set_prepare}

For the ground truth commonsense triples from the CSKBs, we split them into train, development, and test set with the proportion 
8:1:1.  
Negative examples are sampled by selecting a random head and a random tail from the aligned $\mathcal{G}^c$ such that the ratio of negative and ground truth triples is 1:1.
To form a diverse evaluation set, we sample 20K triples from the original automatically constructed test set \finalcopy{(denoted as ``\textit{Original Test Set}'')}, 
20K from the edges in ASER where heads come from CSKBs and tails are from ASER \finalcopy{ (denoted as ``\textit{CSKB head + ASER tail}'')}, 
and 20K triples in ASER where both heads and tails come from ASER \finalcopy{(denoted as ``\textit{ASER edges}'')}. 
The detailed methods of selecting candidate triples for annotation is listed in the Appendix 
\finalcopy{~\ref{sec:appendix_test_set}}.
The distribution of different relations in this evaluation set is the same as in the original test set.
The sampled evaluation set is then annotated to acquire ground labels.

\begin{table}[t]
\small
\centering
\begin{tabular}{p{1.5cm}|ccc}
\toprule
\scriptsize{Relation} & \scriptsize{ATOMIC($_{20}^{20}$)} & \scriptsize{ConceptNet} & \scriptsize{GLUCOSE} \\
\midrule
\scriptsize{\texttt{oEffect}} & 21,497 & 0 & 7,595 \\
\scriptsize{\texttt{xEffect}} & 61,021 & 0 & 30,596 \\
\scriptsize{\texttt{gEffect}} & 0 & 0 & 8,577 \\
\scriptsize{\texttt{oWant}} & 35,477 & 0 & 1,766 \\
\scriptsize{\texttt{xWant}} & 83,776 & 0 & 11,439 \\
\scriptsize{\texttt{gWant}} & 0 & 0 & 5,138 \\
\scriptsize{\texttt{oReact}} & 21,110 & 0 & 3,077 \\
\scriptsize{\texttt{xReact}} & 50,535 & 0 & 13,203 \\
\scriptsize{\texttt{gReact}} & 0 & 0 & 2,683 \\
\scriptsize{\texttt{xAttr}} & 89,337 & 0 & 7,664 \\
\scriptsize{\texttt{xNeed}} & 61,487 & 0 & 0 \\
\scriptsize{\texttt{xIntent}} & 29,034 & 0 & 8,292 \\
\scriptsize{\texttt{isBefore}} & 18,798 & 0 & 0 \\
\scriptsize{\texttt{isAfter}} & 18,600 & 0 & 0 \\
\scriptsize{\texttt{HinderedBy}} & 87,580 & 0 & 0 \\
\scriptsize{\texttt{xReason}} & 189 & 0 & 0 \\
\scriptsize{\texttt{Causes}} & 0 & 42 & 26,746 \\
\scriptsize{\texttt{HasSubEvent}} & 0 & 9,934 & 0 \\
\midrule
\scriptsize{Total}& 578,252 & 10,165 & 126,776 \\

\bottomrule
\end{tabular}
\caption{Relation distribution statistics for different CSKBs. 
Due to the filter in Section~\ref{sec:select_cskb}, the statistics are different from the original papers.
} \label{table:cskb_rel_stat}
\end{table}

\section{Human Annotation}

\subsection{Setups}

The human annotation is carried out on Amazon Mechanical Turk. Workers are provided with sentences in the form of natural language translated from knowledge triples (e.g., for \texttt{xReact}, an ($h$, $r$, $t$) triple is translated to ``If $h$, then, PersonX feels $t$''). 
Additionally, following \citet{hwang2020comet}, annotators are asked to rate each triple in a four-point Likert scale: \textit{Always/Often}, \textit{Sometimes/Likely}, \textit{Farfetched/Never}, and \textit{Invalid}. 
Triples receiving the former two labels will be treated as \textit{Plausible} or otherwise \textit{Implausible}. Each HIT (task) includes 10 triples with the same relation type, and each sentence is labeled by 5 workers. We take the majority vote among 5 votes as the final result for each triple. To avoid ambiguity and control the quality, we finalize the dataset by selecting triples where workers reach an agreement on at least 4 votes.

\subsection{Quality Control}\label{sec:quality_control}
For strict quality control, we carry out two rounds of qualification tests to select workers and provide a special training round. First, workers satisfying the following requirements are invited to participate in our qualification tests: 1) at least 1K HITs approved, and 2) at least 95\% approval rate.
Second, a qualification question set including both straightforward and tricky questions is created by experts, who are authors of this paper and have a clear understanding of this task. 760 triples sampled from the original dataset are annotated by the experts.
Each worker needs to answer a HIT containing 10 questions from the qualification set and their answers are compared with the expert annotation. Annotators who correctly answer at least 8 out of 10 questions are selected in the second round.
671 workers participated in the qualification test, among which 141 (21.01\%) workers are selected as our main round annotators. 
To further enhance the quality, we carry out an extra training round for the main round annotators. For each relation, annotators are asked to rate 10 tricky triples carefully selected by experts. A grading report with detailed explanations on every triple is sent to all workers afterward to help them fully understand the annotation task.

After filtering, we acquire human-annotated labels for 31,731 triples.
The IAA score is 71.51\% calculated using pairwise agreement proportion, and the Fleiss's $\kappa$~\cite{fleiss1971measuring} is 0.43. We further split the proportion of the development set and test set as 2:8. The overall statistics of this evaluation set are presented in Table~\ref{table:annotation_stat}. 
To acquire human performance, we sample 5\% of the triples from the test set, and ask experts as introduced above to provide two additional votes for the triples. 
The agreement between labels acquired by majority voting and the 5+2 annotation labels is used as the final human performance of this task.

\begin{table}[t]
\small
\centering
\begin{tabular}{l|cc|c}
\toprule
 & Dev & Test & Train\\
\midrule
\# Triples & 6,217 & 25,514 & 1,100,362 \\
\% Plausible & 51.05\% & 51.74\% & -\\
\% Novel Nodes & 67.40\% & 70.01\% & - \\
\bottomrule
\end{tabular}
\caption{Statistics of the annotated evaluation set. \# triples indicates the number of triples in the dataset,  \% Plausible indicates the proportion of plausible triples after majority voting, and \% Novel Nodes is the proportion of nodes that do not appear in the training CSKBs. We also report the scale of the un-annotated training set (including random negative examples) for reference.}
\label{table:annotation_stat}
\end{table}


\section{Experiments}

In this section, we introduce the baselines and our proposed model KG-\textsc{Bert}SAGE for the CSKB population task, as well as the experimental setups.

\subsection{ Model}

The objective of a population model is to determine the plausibility of an $(h, r, t)$ triple, where nodes can frequently be out of the domain of the training set. 
In this sense, transductive methods based on knowledge base embeddings~\cite{malaviya2020commonsense} are not studied here.
We present several ways of encoding triples in an inductive manner.

\noindent \textbf{\textsc{Bert}}. The embeddings of $h$, $r$, $t$ are encoded as the embeddings of the [CLS] tokens after feeding them separately as sentences to \textsc{Bert}. For example, the relation \texttt{xReact} is encoded as the \textsc{Bert} embedding of ``[CLS] xReact [SEP]''.
The embeddings are then concatenated as the final representation of the triple, $[s_h, s_r, s_t]$. 
 
\noindent \textbf{\textsc{Bert}SAGE}. The idea of \textsc{Bert}SAGE~\cite{fang2021discos} is to leverage the neighbor information of nodes through a graph neural network layer for their final embedding. For $h$, denote its \textsc{Bert} embedding as $s_h$, then the final embedding of $h$ is $e_h = [s_h, \sum_{v\in \mathcal{N}(h)} s_v / |\mathcal{N}(h)|]$, where $\mathcal{N}(h)$ is the neighbor function that returns the neighbors of $h$ from $\mathcal{G}$. The final representation of the triple is then $[e_h, s_r, e_t]$.

\noindent \textbf{KG-\textsc{Bert}}. KG-\textsc{Bert}(a)~\cite{yao2019kg} encodes a triple by concatenating the elements in $(h, r, t)$ into a single sentence and encode it with \textsc{Bert}. Specifically, the input is the string concatenation of [CLS], $h$, [SEP], $r$, [SEP], $t$, and [SEP].

\noindent \textbf{KG-\textsc{Bert}SAGE}. As KG-\textsc{Bert} doesn't take into account graph structures directly, we propose to add an additional graph SAmpling and AGgregation layer~\cite{hamilton2017inductive} to better learn the graph structures. Specifically, denoting the embedding of the  $(h, r, t)$ triple by KG-\textsc{Bert} as $\text{{KG-\textsc{Bert}}} (h, r, t)$, the model of KG-\textsc{Bert}SAGE is the concatenation of $\text{{KG-\textsc{Bert}}}(h, r, t)$, $\sum_{(r', v)\in\mathcal{N}(h)} \text{ {KG-\textsc{Bert}} }(h, r', v)/|\mathcal{N}(h)|$, and $\sum_{(r', v)\in\mathcal{N}(t)} \text{ {KG-\textsc{Bert}} }(v, r', t)/|\mathcal{N}(t)|$. 
\finalcopy{Here, $\mathcal{N}(h)$ returns the neighboring edges of node $h$.}


More details about the models and experimental details are listed in the Appendix Section~\ref{sec:appendix_model}.

\begin{table}[t]
\small
\centering
\begin{tabular}{l|cc}
\toprule
Relation & \#Eval. & \#Train \\
\midrule
\scriptsize{\texttt{xWant}} & 2,605 & 152,634 \\
\scriptsize{\texttt{oWant}} & 999 & 59,688 \\
\scriptsize{\texttt{gWant}} & 207 & 8,093 \\
\scriptsize{\texttt{xEffect}} & 2,757 & 144,799 \\
\scriptsize{\texttt{oEffect}} & 667 & 46,555 \\
\scriptsize{\texttt{gEffect}} & 287 & 13,529 \\
\scriptsize{\texttt{xReact}} & 2,999 & 100,853 \\
\scriptsize{\texttt{oReact}} & 921 & 38,581 \\
\scriptsize{\texttt{gReact}} & 164 & 4,169 \\
\scriptsize{\texttt{xAttr}} & 2,561 & 152,949 \\
\scriptsize{\texttt{xIntent}} & 1,017 & 59,138 \\
\scriptsize{\texttt{xNeed}} & 1,532 & 98,830 \\
\scriptsize{\texttt{Causes}} & 1,422 & 40,450 \\
\scriptsize{\texttt{xReason}} & 16 & 320 \\
\scriptsize{\texttt{isBefore}} & 879 & 27,784 \\
\scriptsize{\texttt{isAfter}} & 1,152 & 27,414 \\
\scriptsize{\texttt{HinderedBy}} & 4,870 & 127,320 \\
\scriptsize{\texttt{HasSubEvent}} & 459 & 16,410 \\
\bottomrule
\end{tabular}
\caption{Number of triples of each relation in the Eval. (dev+test)  and Train set. 
}
\label{table:relation_group}
\end{table}

\begin{table*}[t]
\scriptsize
\centering
\setlength{\tabcolsep}{0.1em}
\renewcommand{\arraystretch}{1.2}
\begin{tabular}{l|p{0.65cm}<{\centering}p{0.65cm}<{\centering}p{0.65cm}<{\centering}p{0.7cm}<{\centering}p{0.7cm}<{\centering}p{0.7cm}<{\centering}p{0.65cm}<{\centering}p{0.65cm}<{\centering}p{0.65cm}<{\centering}p{0.55cm}<{\centering}p{0.66cm}<{\centering}p{0.65cm}<{\centering}p{0.65cm}<{\centering}p{0.8cm}<{\centering}p{0.78cm}<{\centering}p{0.7cm}<{\centering}p{0.7cm}<{\centering}p{0.95cm}<{\centering}| p{0.7cm}<{\centering}}
\toprule
 \ Relation & \tiny{\texttt{xWnt}}&\tiny{\texttt{oWnt}}&\tiny{\texttt{gWnt}}&\tiny{\texttt{xEfct}}&\tiny{\texttt{oEfct}}&\tiny{\texttt{gEfct}}&\tiny{\texttt{xRct}}&\tiny{\texttt{oRct}}&\tiny{\texttt{gRct}}&\tiny{\texttt{xAttr}}&\tiny{\texttt{xInt}}&\tiny{\texttt{xNeed}}&\tiny{\texttt{Cause}}&\tiny{\texttt{xRsn}}&\tiny{\texttt{isBfr}}&\tiny{\texttt{isAft}}&\tiny{\texttt{Hndr}}.&\tiny{\texttt{HasSubE}}.&\ all\\
 \midrule
 \textsc{\ Bert} & 57.7 & 64.9 & 66.3 & 59.1 & 66.2 & 60.0 & 50.6 & \textbf{68.7} & 72.3 & 56.2 & 63.9 & 56.4 & 48.3 & 34.5 & 59.2 & 58.0 & 66.1 & 73.0 & 59.4 \\
 \textsc{\ BertSAGE} & 54.7& 58.9& 58.0& 58.0& 70.0& 54.7& 52.8& 62.4& \textbf{76.6}& 55.0& 61.0& 57.1& 46.2& 45.5& 66.7& 64.9& 69.6& \textbf{80.4} & 60.0 \\
 \textsc{\ KG-Bert} & 63.2 & \textbf{69.8} & \textbf{69.0} & 68.0 & 70.6 & 61.0 & 57.0 & 64.0 & 73.8 & \textbf{59.5} & \textbf{64.9} & 64.6 & 47.4 & \textbf{90.9} & 78.0 & \textbf{77.5} & 75.9 & 68.5 & 66.1 \\
 \textsc{\ KG-BertSAGE \quad }& \textbf{66.0} & 68.9 & 68.6 & \textbf{68.2} & \textbf{70.8}& \textbf{62.3} & \textbf{60.5} & 64.6 & 74.1 & 59.1 & 63.0 & \textbf{65.4} &\textbf{50.0} & 76.4 & \textbf{78.2} & 77.4 & \textbf{77.5} & 67.0 & \textbf{67.2}\\
 \midrule
 Human & 86.2 & 86.8 & 83.3 & 85.2 & 83.9 & 79.8 & 81.1 & 82.6 & 76.5 & 82.6 & 85.6 & 87.4 & 80.1 & 73.7 & 89.8 & 89.9 & 85.3 & 85.7 & 84.4\\
\bottomrule
\end{tabular}
\caption{Experimental results on CSKB population. We report the AUC ($\times$100) here for each relation. The improvement under ``all'' is statistically significant using Randomization Test~\cite{cohen1995empirical}, with $p<0.05$.}
\label{table:result_main_each_rel}
\end{table*}


\begin{table*}[h]
\scriptsize
\centering
\setlength{\tabcolsep}{0.1em}
\renewcommand{\arraystretch}{1.2}
\begin{tabular}{l|p{0.65cm}<{\centering}p{0.65cm}<{\centering}p{0.65cm}<{\centering}p{0.7cm}<{\centering}p{0.7cm}<{\centering}p{0.7cm}<{\centering}p{0.65cm}<{\centering}p{0.65cm}<{\centering}p{0.65cm}<{\centering}p{0.55cm}<{\centering}p{0.66cm}<{\centering}p{0.65cm}<{\centering}p{0.65cm}<{\centering}p{0.8cm}<{\centering}p{0.78cm}<{\centering}p{0.7cm}<{\centering}p{0.7cm}<{\centering}p{0.95cm}<{\centering}| p{0.6cm}<{\centering}}

\toprule
\ Relation & \tiny{\texttt{xWnt}}&\tiny{\texttt{oWnt}}&\tiny{\texttt{gWnt}}&\tiny{\texttt{xEfct}}&\tiny{\texttt{oEfct}}&\tiny{\texttt{gEfct}}&\tiny{\texttt{xRct}}&\tiny{\texttt{oRct}}&\tiny{\texttt{gRct}}&\tiny{\texttt{xAttr}}&\tiny{\texttt{xInt}}&\tiny{\texttt{xNeed}}&\tiny{\texttt{Cause}}&\tiny{\texttt{xRsn}}&\tiny{\texttt{isBfr}}&\tiny{\texttt{isAft}}&\tiny{\texttt{Hndr}}.&\tiny{\texttt{HasSubE}}.&\ all\\

 \midrule
 KG-\textsc{Bert} \\
 $\cdot$ on \scriptsize{ATOMIC($_{20}^{20}$)} & 61.0& 64.2& 68.0& 62.9& 67.1& \textbf{64.8}& \textbf{58.8}& 60.2& 68.6& 58.9& 62.4& 63.7& \textbf{55.8}& 58.2& 77.7& 76.7& 75.5& 67.6 &65.2 \\
 $\cdot$ on GLUCOSE & 62.3& 67.6& \textbf{69.2}& 61.6& \textbf{71.5}& 57.3& 58.0& 63.4& \textbf{77.0}& 57.7& 61.0& 50.4& 48.1& 72.7& 61.0& 50.6& 59.2& 68.0 & 59.2 \\
 $\cdot$ on ConceptNet & 58.0& 62.0& 59.4& 56.2& 52.5& 61.4& 52.3& 57.0& 54.4& 57.1& 61.8& 57.4& 55.6& 78.2& 61.8& 60.8& 63.2& 60.9 & 58.3 \\
 $\cdot$ on all & \textbf{63.2} & \textbf{69.8} & 69.0 & \textbf{68.0} & 70.6 & 61.0 & 57.0 & \textbf{64.0} & 73.8 & \textbf{59.5} & \textbf{64.9} & \textbf{64.6} & 47.4 & \textbf{90.9} & \textbf{78.0} & \textbf{77.5} & \textbf{75.9} & \textbf{68.5} & \textbf{66.1} \\
 \midrule
 KG-\textsc{Bert}SAGE  \\
 $\cdot$ on \scriptsize{ATOMIC($_{20}^{20}$)} & 63.1& 64.7& 65.6& 63.7& 67.5& \textbf{65.7}& 56.1& 60.3& 64.9& 56.8& 60.5& 63.7& \textbf{56.5}& 65.5& 76.9& 76.6& 76.9& 63.8 & 65.1   \\
 $\cdot$ on GLUCOSE & 61.7& 68.3& \textbf{70.8}& 61.1& \textbf{71.9}& 60.1& 56.1& 61.4& 71.3& 56.5& 60.5& 46.8& 50.5& 69.1& 60.6& 51.7& 60.0& \textbf{72.4} & 58.9  \\
 $\cdot$ on ConceptNet & 57.7& 55.0& 59.8& 60.1& 57.3& 62.2& 50.2& 50.9& 50.9& 52.3& 56.8& 52.1& 52.6& 70.9& 53.8& 44.5& 58.3& 59.8 & 55.0\\
 $\cdot$ on all & \textbf{66.0} & \textbf{68.9} & 68.6 & \textbf{68.2} & 70.8 & 62.3 & \textbf{60.5} & \textbf{64.6} & \textbf{74.1} & \textbf{59.1} & \textbf{63.0} & \textbf{65.4} & 50.0 & \textbf{76.4} & \textbf{78.2} & \textbf{77.4} & \textbf{77.5} & 67.0 & \textbf{67.2} \\
\bottomrule
\end{tabular}
\caption{Effects of different training sets. }\label{table:result_dataset_ablation}
\end{table*}


\subsection{Setup}

We train the population model using a triple classification task, where ground truth triples come from the original CSKB, and the negative examples are randomly sampled from the aligned graph $\mathcal{G}^c$ . The model needs to discriminate whether an $(h, r, t)$ triple in the human-annotated evaluation set is plausible or not. 
For evaluation, we use the AUC score as the evaluation metric, as this commonsense reasoning task is essentially a ranking task that is expected to rank plausible assertions higher than those farfetched assertions.

We use \textsc{Bert}$_{base}$ from the Transformer\footnote{https://transformer.huggingface.co/} library, and use learning rate $5\times 10^{-5}$ and batch size $32$ for all models.
\finalcopy{The statistics of each relation is shown in Table~\ref{table:relation_group}.}
We select the best models individually for each relation based on the corresponding development set.
Besides AUC scores for each relation, we also report the AUC score for all relations by the weighted sum of the break-down scores, weighted by the proportion of test examples of the relation. This is reasonable as AUC essentially represents the probability that a positive example will be ranked higher than a negative example.

\subsection{Main Results}

The main experimental results are shown in Table~\ref{table:result_main_each_rel}. KG-\textsc{Bert}SAGE performs the best among all, as it both encodes an $(h, r, t)$ as a whole and takes full advantage of neighboring information in the graph. 
Moreover, all models are significantly lower than human performance with a relatively large margin.

ASER can on the one hand provide candidate triples for populating CSKBs, and can on the other hand provide graph structure for learning commonsense reasoning. 
From the average degree in Table~\ref{table:overall_match_coverage}, the graph acquired by grounding CSKBs to ASER can provide far more
neighbor information than using the CSKBs only.
While KG-\textsc{Bert} treats the task directly as a simple triple classification task and takes only the triples as input, it does not explicitly take into consideration the graph structure. KG-\textsc{Bert}SAGE on the other hand leverages an additional GraphSAGE layer to aggregate the graph information from ASER, thus achieving better performance. It demonstrates that it is beneficial to incorporate those un-annotated ASER graph structures where multiple-hop paths are grounded between commonsense heads and tails. 
Though \textsc{BertSAGE} also incorporates neighboring information, it only leverages the ASER nodes representation and ignores the complete relational information of triples as KG-\textsc{Bert}SAGE does. As a result, it doesn't outperform \textsc{Bert} by much for the task.

\subsection{Zero-shot Setting} 

We also investigate the effects of different training CSKBs as shown in Table~\ref{table:result_dataset_ablation}. Models are then trained on the graphs only consisting of commonsense knowledge from ATOMIC($_{20}^{20}$),
GLUCOSE, and ConceptNet, respectively. The models trained on all CSKBs achieve better performance both for each individual relation and on the whole. We can conclude that more high-quality commonsense triples for training from diverse dimensions can benefit the performance of such commonsense reasoning. 

When trained on each CSKB dataset, there are some relations that are never seen in the training set. 
As all of the models use \textsc{Bert} to encode relations, the models are \textit{inductive} and can thus reason triples for unseen relations in a zero-shot setting.
For example, the \texttt{isBefore} and \texttt{isAfter} relations are not presented in GLUCOSE, while after training KG-\textsc{Bert}SAGE on GLUCOSE, it can still achieve fair AUC scores. Though not trained explicitly on the \texttt{isBefore} and \texttt{isAfter}  relations, the model can transfer the knowledge from other relations and apply them to the unseen ones. 


\section{Error Analysis}

\begin{table}[t]
\small
\centering
\begin{tabular}{l|ccc}
\toprule
Model & \tabincell{c}{\textit{Original}\\ \textit{Test Set}} & \tabincell{c}{\textit{CSKB head }\\ \textit{+ ASER tail}} & \tabincell{c}{\textit{ASER}\\ \textit{edges}} \\
\midrule
\textsc{Bert} & 65.0 & 47.9 & 44.6 \\
\textsc{BertSAGE} & 67.2 & 49.4 & 46.2 \\
\textsc{KG-Bert} & 77.8 & 55.2 & 50.3\\
\textsc{KG-BertSAGE} & \textbf{78.2} & \textbf{57.5} & \textbf{52.3} \\
\bottomrule
\end{tabular}
\caption{AUC scores grouped by the types of the evaluation sets defined in~\ref{section:eval_set_prepare}. The latter two groups are harder for neural models to distinguish.}
\label{table:results_by_class}
\end{table}

\finalcopy{As defined in Section~\ref{section:eval_set_prepare}, the evaluation set is composed of three parts, edges coming from the original test set (\textit{Original Test Set}), edges where heads come from CSKBs and tails from ASER (\textit{CSKB head + ASER tail}), and edges from the whole ASER graph (\textit{ASER edges}). The break-down AUC scores of different groups given all models are shown in Table~\ref{table:results_by_class}. The performances under the \textit{Original Test Set} of all models are remarkably better than the other two groups, as the edges in the original test set are from the same domain as the training examples. The other two groups, where there are more unseen nodes and edges, are harder for the neural models to distinguish. 
The results show that simple commonsense reasoning models studied in this paper struggle to be generalized to unseen nodes and edges. 
As a result, in order to improve the performance of this CSKB population task, more attention should be paid to the generalization ability of commonsense reasoning on unseen nodes and edges.
}

\finalcopy{
Moreover, by taking a brief inspection of the test set, we found that errors occur when encountering triples that are not logically sound but semantically related. 
Some examples are presented in Table~\ref{table:error_case}. For the triple (\textit{PersonX} go to nurse, \texttt{xEffect}, \textit{PersonX} use to get headache), the head event and tail event are highly related. However, the fact that someone gets a headache should be the reason instead of the result of going to the nurse. More similar errors are presented in the rest of the table.
These failures may be because when using \textsc{Bert}-based models the training may not be well performed for the logical relations or discourse but still recognizing the semantic relatedness patterns.
}

\begin{table}[t]
\centering
\tiny
\setlength{\tabcolsep}{0.1em}
\begin{tabular}{p{0.14\textwidth}|p{0.07\textwidth}<{\centering}|p{0.18\textwidth}|p{0.03\textwidth}<{\centering}|p{0.03\textwidth}<{\centering}}
\toprule
Head & Relation & Tail & Label & Pred.\\
\midrule
\textit{PersonX} go to nurse & \texttt{xEffect} & \textit{PersonX} use to get headache & 0 & 1\\
\textit{PersonX} have a quiz & \texttt{Causes} & \textit{PersonX} have pen & 0 & 1 \\
\textit{PersonX} be strong & \texttt{oWant} & \textit{PersonY} like \textit{PersonX} & 0 & 1\\
\textit{PersonX} feel a pain & \texttt{xIntent} & PersonX finger have be chop off & 0 & 1\\
 \bottomrule
 
\end{tabular}%

\caption{ \finalcopy{Examples of error predictions made by \textsc{KG-BertSAGE}, where the head and tail are semantically related while not conformed to the designated commonsense relation. }} \label{table:error_case}
\end{table}

\section{Conclusion}
In this paper, we benchmark the CSKB population task by proposing a dataset by aligning four popular CSKBs and an eventuality graph ASER, and provide a high-quality human-annotated evaluation set to test models' reasoning ability. We also propose KG-\textsc{Bert}SAGE to both incorporate the semantic of knowledge triples and the subgraph structure to conduct reasoning, which achieves the best performance among other counterparts. Experimental results also show that the task of reasoning unseen triples outside of the domain of CSKB is a hard task 
where current models are far away from human performance, which brings challenges to the community for future research.

\section*{Acknowledgement}
The authors of this paper were supported by the NSFC Fund (U20B2053) from the NSFC of China, the RIF (R6020-19 and R6021-20) and the GRF (16211520) from RGC of Hong Kong, the MHKJFS (MHP/001/19) from ITC of Hong Kong, with special thanks to the Gift Fund from Huawei Noah’s Ark Lab.



\bibliography{emnlp2021}
\bibliographystyle{acl_natbib}

\clearpage

\appendix

\section{Additional Details of Commonsense Relations}

During human annotation, we translate the symbolic knowledge triples into human language for annotators to better understand the questions. A $(h, r, t)$ triple where $h$, $r$, and $t$ are the head, relation, and tail, is translated to \textit{if $h$, then [Description], $t$}. Here, the description placeholder \textit{[Description]} comes from rules in Table~\ref{table:cskb_rel_def}, which is modified from~\citet{hwang2020comet}. These descriptions can also be regarded as definitions of those commonsense relations.

Moreover, the definitions of the discourse relations in ASER are presented in Table~\ref{table:aser_rel_def}. We also present the statistics of relation distribution for ASER$_{norm}$ in Table~\ref{table:aser_rel_stat}. 

\section{Additional Details of Pre-processing}

\subsection{Examples of Format Unification}

Table~\ref{table:alignment_example} demonstrates several examples for unifying the formats of different resources.  In ConceptNet and Knowlywood, the nodes are mostly \textit{verb} or \textit{verb-object} phrases, and we add a subject ``\textit{PersonX}'' in front of each node. For ATOMIC, the main modification part is the tails, where subjects tend to be missing. We treat \textit{agent}-driven (relations investigating causes and effects on \textit{PersonX}) and \textit{theme}-driven (relations investigating causes and effects on \textit{PersonY}) differently, and add \textit{PersonX} or \textit{PersonY} in front of the tails whose subjects are missing. For ASER, rules are used to discriminate \textit{PersonX} and \textit{PersonY} in a certain edge. Two examples for ASER and ATOMIC demonstrating the differences between \textit{PersonX} and \textit{PersonY}  are provided in the table. For GLUCOSE, we simply replace \textit{SomeoneA} with \textit{PersonX} and \textit{SomeoneB} with \textit{PersonY} accordingly. Moreover, all the words are lemmatized using Stanford CoreNLP parser\footnote{https://stanfordnlp.github.io/CoreNLP/} to normalized forms.

\begin{table}[t]
\scriptsize
\centering
\begin{tabular}{p{1.5cm}|l}
\toprule
Relation & Decriptions\\
\midrule
\scriptsize{\texttt{oEffect}} & then, \textit{PersonY} will \\
\scriptsize{\texttt{xEffect}} & then, \textit{PersonX} will\\
\scriptsize{\texttt{gEffect}} & then, other people or things will\\
\scriptsize{\texttt{oWant}} &  then, \textit{PersonY} wants to \\
\scriptsize{\texttt{xWant}} & then, \textit{PersonX} wants to\\
\scriptsize{\texttt{gWant}} & then, other people or things want to\\
\scriptsize{\texttt{oReact}} & then, \textit{PersonY} feels\\
\scriptsize{\texttt{xReact}} & then, \textit{PersonX} feels\\
\scriptsize{\texttt{gReact}} & then, other people or things feel\\
\scriptsize{\texttt{xAttr}} & \textit{PersonX} is seen as\\
\scriptsize{\texttt{xNeed}} & but before, \textit{PersonX} needed \\
\scriptsize{\texttt{xIntent}} & because \textit{PersonX} wanted \\
\scriptsize{\texttt{isBefore}} & happens before \\
\scriptsize{\texttt{isAfter}} & happens after\\
\scriptsize{\texttt{HinderedBy}} & can be hindered by \\
\scriptsize{\texttt{xReason}} & because\\
\scriptsize{\texttt{Causes}} & causes \\
\scriptsize{\texttt{HasSubEvent}} & includes the event/action \\

\bottomrule
\end{tabular}
\caption{Descriptions of different commonsense relations, which are translation rules from knowledge triples $(h, r, t)$ to human language, ``\textit{if h, then [Description], t}''~\cite{hwang2020comet}.} \label{table:cskb_rel_def}
\end{table}

\begin{table}[t]
\scriptsize
\centering
\begin{tabular}{p{2.4cm}|p{4.5cm}}
\toprule
Relation & Decriptions\\
\midrule
\scriptsize{\texttt{Precedence}} & $h$ happens before $t$  \\
\scriptsize{\texttt{Succession}} & $h$ happens after $h$ \\
\scriptsize{\texttt{Synchronous}} & $h$ happens the same time as $t$ \\
\scriptsize{\texttt{Reason}} & $h$ happens because $t$ \\
\scriptsize{\texttt{Result}} & $h$ result in $t$ \\
\scriptsize{\texttt{Condition}} & Only when $t$ happens, $h$ can happen \\
\scriptsize{\texttt{Contrast}} & $h$ and $t$ share significant difference regarding some property\\
\scriptsize{\texttt{Concession}} & $h$ and $t$ result in another opposite event \\
\scriptsize{\texttt{Alternative}} & $h$ and $t$ are alternative situations of each other.  \\
\scriptsize{\texttt{Conjunction}} & $h$ and $t$ both happen \\
\scriptsize{\texttt{Restatement}} & $h$ restates $t$  \\
\scriptsize{\texttt{Instantiation}} & $t$ is a more detailed description of $h$ \\
\scriptsize{\texttt{ChosenAlternative}} & $h$ and $t$ are alternative situations of each other, but the subject prefers $h$ \\
\scriptsize{\texttt{Exception}} & $t$ is an exception of $h$  \\
\scriptsize{\texttt{Co\_Occurrence}} & $h$ and $t$ co-occur at the same sentence \\

\bottomrule
\end{tabular}
\caption{Descriptions of discourse relations in ASER~\cite{zhang2021aser}.} \label{table:aser_rel_def}
\end{table}

\begin{table}[t]
\scriptsize
\centering
\begin{tabular}{l|c}
\toprule
Relation & number of edges \\
\midrule
\scriptsize{\texttt{Precedence}} & 4,957,481 \\
\scriptsize{\texttt{Succession}} & 1,783,154 \\
\scriptsize{\texttt{Synchronous}} & 8,317,572 \\
\scriptsize{\texttt{Reason}} & 5,888,968 \\
\scriptsize{\texttt{Result}} & 5,562,565 \\
\scriptsize{\texttt{Condition}} & 8,109,020 \\
\scriptsize{\texttt{Contrast}} & 23,208,195 \\
\scriptsize{\texttt{Concession}} & 1,189,167 \\
\scriptsize{\texttt{Alternative}} & 1,508,729 \\
\scriptsize{\texttt{Conjunction}} & 37,802,734 \\
\scriptsize{\texttt{Restatement}} & 159,667 \\
\scriptsize{\texttt{Instantiation}} & 33,840 \\
\scriptsize{\texttt{ChosenAlternative}} & 91,286 \\
\scriptsize{\texttt{Exception}} & 51,502 \\
\scriptsize{\texttt{Co\_Occurrence}} & 124,330,714 \\
\midrule
 \scriptsize{Total} & 222,994,594 \\
\bottomrule
\end{tabular}
\caption{Statistics of relations in ASER$_{norm}$. 
}\label{table:aser_rel_stat}
\end{table}

\begin{table*}[t]
    \scriptsize
    \centering
    \renewcommand\arraystretch{1.5}
    \begin{tabular}{l|lll|ll}
    \toprule
    \multirow{2}{*}{Resource}  & \multicolumn{3}{c|}{Original Format} & \multicolumn{2}{c}{Aligned Format}\\
    ~ & \multicolumn{1}{c}{Head} & \multicolumn{1}{c}{Relation} &
    \multicolumn{1}{c|}{Tail} & \multicolumn{1}{c}{Head} & \multicolumn{1}{c}{Tail}  \\
    \midrule
    ConceptNet & get exercise & \scriptsize{\texttt{HasSubEvent}} & ride bicycle & \textit{PersonX} get exercise & \textit{PersonX} ride bicycle \\
    \hline
    \multirow{2}{*}{ATOMIC($_{20}^{20}$)} & \textit{PersonX} gets exercise &  \scriptsize{\texttt{xReact}} & tired & \textit{PersonX} get exercise & \textit{PersonX} be tired \\
    ~ & \textit{PersonX} visits \textit{PersonY} at work & \texttt{oEffect} & say hello & \textit{PersonX} visits \textit{PersonY} & \textit{PersonY} say hello\\
    \hline
    GLUCOSE & \textit{SomeoneA} gets exercise & \scriptsize{\texttt{Dim 1 (xEffect)}}  & \textit{SomeoneA} gets tired & \textit{PersonX} get exercise & \textit{PersonX} be tired \\
    \hline
    Knowlywood & get exercise & \texttt{NextActivity} & take shower & \textit{PersonX} get exercise & \textit{PersonX} take shower \\
    \hline
    \multirow{2}{*}{ASER} & he gets exercise & \texttt{Result} & he is tired & \textit{PersonX} get exercise & \textit{PersonX} be tired \\
    ~ & he visits her at work & \texttt{Precedence} & she is happy & \textit{PersonX} visit \textit{PersonY} at work & \textit{PersonY} is happy\\
    \bottomrule
    \end{tabular}
    \caption{Examples of format unification of CSKBs and eventuality graphs.} \label{table:alignment_example}
\end{table*}

\begin{table*}[t]
    \scriptsize
    \renewcommand\arraystretch{1.5}
    \begin{tabular}{m{0.18\textwidth}|m{0.6\textwidth}<{\centering}|m{0.15\textwidth}}
    \toprule
    \tabincell{l}{Commonsense\\Relations} & ASER Relations & Patterns \\
    \midrule
     \tabincell{l}{\texttt{Effect}, \texttt{Want}\\\texttt{isBefore}, \texttt{Causes}}  & \texttt{Result}, \texttt{Precedence}, \texttt{Condition}$^{-1}$, \texttt{Succession}$^{-1}$, \texttt{Reason}$^{-1}$ & - \\
    \texttt{React} & \texttt{Result}, \texttt{Precedence}, \texttt{Condition}$^{-1}$, \texttt{Succession}$^{-1}$, \texttt{Reason}$^{-1}$ & \textit{s-v/be-a/o}, \textit{s-v-be-a/o, s-v, spass-v}   \\
    \texttt{xIntent, xNeed, isAfter} &\texttt{Condition, Succession, Reason, Result$^{-1}$, Precedence$^{-1}$ }& - \\
    \texttt{xAttr} &  \texttt{Synchronous$^{\pm 1}$, Reason$^{\pm1}$, Result$^{\pm 1}$, Condition$^{\pm 1}$, Conjunction$^{\pm 1}$, Restatement$^{\pm 1}$} & \textit{s-be-a/o, s-v-a, s-v-be-a/o, s-v, spass-v} \\
    \texttt{HinderedBy} & \texttt{Concession, Alternative} & - \\
    \texttt{HasSubEvent}& \texttt{Synchronous$^{\pm 1}$, Conjunction$^{\pm 1}$} & - \\
    \bottomrule
    \end{tabular}
    \caption{Rules of selecting candidate triples. \finalcopy{For a certain commonsense relation \texttt{r$_{cs}$} in the first column, (\textit{head}, \texttt{r$_{ASER}$}, \textit{tail}) in ASER, where \texttt{r$_{ASER}$} belongs to the corresponding cell in the second column, can be selected as a candidate (\textit{head}, \texttt{r$_{cs}$}, \textit{tail}) for annotation.}}
     \label{table:aser_candidate}
\end{table*}

\begin{table}[h]
\small
\centering
\begin{tabular}{l|c}
\toprule
Model & Average AUC \\
\midrule
KG-\textsc{Bert}SAGE (Dir) & 66.2 \\
KG-\textsc{Bert}SAGE (Undir) & \textbf{67.2} \\
\bottomrule
\end{tabular}
\caption{Experimental results using two different neighboring functions.}
\label{table:result_neighboring_func}
\end{table}

\begin{table*}[t]
    \scriptsize
    \centering
    \renewcommand\arraystretch{1.5}
    \begin{tabular}{l|l|l|l|l}
    \toprule
        Head & Relation & Tail & Label & Source\\
        \midrule
                \textit{PersonX} give \textit{PersonY} ride & \texttt{xNeed} & \textit{PersonX} need to wear proper clothes & Plau. & \multirow{3}{*}{\tabincell{c}{Triples in CSKBs\\ (\textit{Original Test Set}) }} \\
        \textit{PersonX} be wait for taxi & \texttt{isAfter} & \textit{PersonX} hail a taxi & Plau. & \\
        \textit{PersonX} be diagnose with something & \texttt{Causes} & \textit{PersonX} be sad & Plau. & \\
        \textit{PersonX} feel something	& \texttt{xEffect} &	\textit{PersonX} figure & Implau. &  \multirow{3}{*}{Randomly sampled examples}\\
        \textit{PersonX} be patient with ignorance	& \texttt{HinderedBy} &	\textit{PersonY} have the right vocabulary & Implau. & \\
        \textit{PersonY} grasp \textit{PersonY} meaning	& \texttt{HasSubEvent} &	\textit{PersonY} open it mechanically & Implau.&  \\
        \midrule
        \textit{PersonX} spill coffee	& \texttt{oEffect} &	\textit{PersonY} have to server & Plau. & \multirow{6}{*}{\textit{CSKB head + ASER tail}} \\
        \textit{PersonX} care for \textit{PersonY}	& \texttt{xNeed} &	\textit{PersonX} want to stay together & Plau. & \\
        \textit{PersonX} be save money	& \texttt{HasSubEvent} &	PeopleX can not afford something & Plau. & \\
        \textit{PersonX} decide to order a pizza	& \texttt{xReact} &	\textit{PersonX} have just move & Implau. & ~\\
        it be almost christmas	& \texttt{gReact} &	\textit{PersonX} be panic & Implau. & ~\\
        arm be break	& \texttt{isBefore} &	\textit{PersonY} ask & Implau. & ~\\
        \midrule
        \textit{PersonX} go early in morning &	\texttt{xEffect} &	\textit{PersonX} do not have to deal with crowd & Plau. & \multirow{6}{*}{\textit{ASER edges}} \\
        \textit{PersonX} have take time to think it over  \textit{PersonX}	& \texttt{xReact} & 	\textit{PersonX} be glad & Plau. & \\
        \textit{PersonX} have a good work-life balance & 	\texttt{xIntent} & 	\textit{PersonX} be happy & Plau. & \\
        \textit{PersonX} weight it by value	& \texttt{oWant}	& \textit{PersonY} bet & Implau. & \\
        \textit{PersonX} be hang out on reddit	& \texttt{oReact}	& \textit{PersonY} can not imagine & Implau. & \\
        \textit{PersonX} can get \textit{PersonY} out shell	& \texttt{xIntent}	& \textit{PersonX} just start poach \textit{PersonY} & Implau. & \\

    \bottomrule
    \end{tabular}
    \caption{Examples of the human-annotated populated triples.} \label{table:population_example}
\end{table*}

\subsection{Selecting Candidate Triples from ASER} \label{sec:appendix_test_set}

\finalcopy{The evaluation set comes from three parts:
\begin{enumerate}
    \item \textit{Original Test Set:} The edges that are randomly sampled from the original automatically constructed test set, as illustrated in Section~\ref{section:eval_set_prepare}.
    \item \textit{CSKB head + ASER tail:} The edges are sampled from the edges in ASER where the heads come from the nodes in CSKBs and tails from ASER.  This corresponds to the settings in COMET~\cite{bosselut2019comet} and DISCOS~\cite{fang2021discos}.
    \item \textit{ASER edges:} The edges are sampled from the whole ASER graph. 
\end{enumerate}
}

Instead of randomly sampling negative examples which may be easy to distinguish, we sample some candidate edges from ASER with some simple rules to fit the chronological order and syntactical patterns for each commonsense relation, thus providing a harder evaluation set for machines to concentrate more on commonsense. 
The discourse relations defined in ASER at Table~\ref{table:aser_rel_def} inherently represent some chronological order, which can be matched to each commonsense relation based on some alignment rules. 

First, for each commonsense relation, we sample the edges in ASER with the same basic chronological and logical meaning. 
For example, the \texttt{Result} relation from ASER, which is a discourse relation where the tail is a result of the head, can be served as a candidate for the \texttt{xEffect} commonsense relation, where a tail is the effect or consequence of the head. 
Alternatively, we can also regard a (\textit{tail}, \texttt{Succession}$^{-1}$, \textit{head}), which is the inversion of (\textit{head}, \texttt{Succession}, \textit{tail}), as a candidate \texttt{xEffect} relation, as in \texttt{Succession}, the head happens after the tail.
By providing candidate triples with the same chronological relation, the models will need to focus more on the subtle commonsense connection within the triple.
Second, we restrict the dependency patterns of the candidate edges.
For the stative commonsense relation such as \texttt{xAttr}, where the tails are defined to be a state, we restrict the tails from ASER to be of patterns such as \textit{s-v-o} and \textit{s-v-a}. 
This also filters out some triples that are obviously false as they are not actually describing a state. 
Detailed selection rules for each commonsense relation are defined in Table~\ref{table:aser_candidate}.

\finalcopy{Besides the above selected edges, we also sample some edges from ASER that are reverse to the designated discourse relations. For example, for the commonsense relation \texttt{xEffect}, the above rules will select discourse edges with patterns like (\textit{head}, \texttt{Result}, \textit{tail}) to constitute a candidate \texttt{xEffect} relation (\textit{head}, \texttt{xEffect}, \textit{tail}). In addition to that, we also sample some edges with reverse relations, like (\textit{tail}, \texttt{Result}, \textit{head}), to form a candidate edge (\textit{head}, \texttt{xEffect}, \textit{tail}), to make the annotated edges more diverse. }

\subsection{Examples of Populated Triples}

Examples of the annotations of the populated triples are listed in Table~\ref{table:population_example}. 
The source of the triples is from the three types defined in Section~\ref{sec:appendix_test_set}.
In the \textit{Original Test Set} category, the triples are composed of two parts, one is the ground truth triples from the original CSKBs, and one is triples randomly sampled from $\mathcal{G}^c$.

\section{Additional Details of the Models}\label{sec:appendix_model}

\subsection{Model Details}

For a $(h, r, t)$ triple, we denote the word tokens of $h$ and $t$ as $w_1^h, w_2^h, \cdots, w_l^h$ and $w_1^t, w_2^t, \cdots, w_m^t$, where $l$ and $m$ are the lengths of the corresponding sentences. For the \textsc{Bert} model, the model takes ``[CLS] $w_1^h$ $w_2^h$ $\cdots$ $w_l^h$ [SEP]'' as the input to a \textsc{Bert}$_{base}$ encoder, and the corresponding embedding for the [CLS] token is regarded as the final embedding $s_h$ of the head $h$. 
The tail $t$ is encoded as $s_t$ similarly with the head. 
For the relation $r$, we feed the name of the relation directly between [CLS] and [SEP] into \textsc{Bert}, which is ``[CLS] $r$ [SEP]'', and we use the corresponding embedding for the [CLS] token as the embedding of $r$ as $s_r$. As \textsc{Bert} adopts sub-word encoding, the relations, despite being complicated symbols, can be split into several meaningful components for \textsc{Bert} to encode. For example, \texttt{xReact} will be split into ``x'' and ``react'', which can demonstrates both the semantics of ``x'' (the relation is based on \textit{PersonX}) and ``react'' (the reaction of the head event).

For \textsc{KG-Bert}, we encode a $(h, r, t)$ triple by feeding the concatenation of the three elements into \textsc{Bert}. Specifically, ``[CLS] $w_1^h$ $w_2^h$ $\cdots$ $w_l^h$ [SEP] $r$ [SEP] $w_1^t$ $w_2^t$ $\cdots$ $w_l^t$  [SEP]'' is fed into \textsc{Bert} and we regard the embedding of [CLS] as the final representation of the triple. 

Denote the embedding of a $(h, r, t)$ triple acquired by \textsc{KG-Bert} as \textsc{KG-Bert}$(h, r, t)$. The function $\mathcal{N}(v)$ is defined as returning the incoming neighbor-relation pairs, which is $\{(r, u) | (u, r, v) \in \mathcal{G} \}$ ($\mathcal{G}$ is ASER in our case.) $\mathcal{N}(v)$ is defined as the function that returns the set $\{(r, u) | (v, r, u) \in \mathcal{G} \}$, which are neighboring edges. The model KG-\textsc{Bert}SAGE then encodes a $(h, r, t)$ triple as:

\begin{align}
    [ &\text{\footnotesize{KG-\textsc{Bert}}}(h, r, t), \nonumber\\
    &\sum_{(r', v)\in\mathcal{N}(h)} \text{\footnotesize{KG-\textsc{Bert}}}(h, r', v)/|\mathcal{N}(h)|, \nonumber\\
    &\sum_{(r', v)\in\mathcal{N}(t)} \text{\footnotesize{KG-\textsc{Bert}}}(v, r', t)/|\mathcal{N}(t)| ] \nonumber
\end{align}

Moreover, as the average number of degrees for nodes in ASER is quite high, we follow the idea in GraphSAGE~\cite{hamilton2017inductive} to conduct uniform sampling on the neighbor set. 
4 neighbors are randomly sampled during training.

\subsection{Neighboring Function $\mathcal{N}$}

\finalcopy{
The edges in ASER are directed. We try two kinds of neighboring functions :
\begin{align}\label{eq:neigh_dir}
  \mathcal{N}(v) = \{(r, u) | (v, r, u) \in \mathcal{G} \}  
\end{align}
\begin{align}\label{eq:neigh_undir}
  \mathcal{N}(v) = \{(r, u) | (v, r, u) \in \mathcal{G} \text{\ or\ } (u, r, v) \in \mathcal{G} \}
\end{align}
Equation~(\ref{eq:neigh_dir}) is the function that returns the outgoing edges of vertex $v$. Equation~(\ref{eq:neigh_undir}) is the function that returns the bi-directional edges of vertex $v$. The overall results using the two mechanisms of \textsc{KG-BertSAGE} is shown in Table~\ref{table:result_neighboring_func}. By incorporating bi-directional information of each vertex, the performance of CSKB population can be largely improved.
}



\end{document}